\documentclass{article}

\PassOptionsToPackage{numbers, compress}{natbib}
\usepackage{jmlr2e}   % at submission time, please omit the final option (anonimisation etc.)

\usepackage[utf8]{inputenc} % allow utf-8 input
\usepackage[T1]{fontenc}    % use 8-bit T1 fonts
\usepackage{hyperref}       % hyperlinks
\usepackage{url}            % simple URL typesetting
\usepackage{booktabs}       % professional-quality tables
\usepackage{amsfonts}       % blackboard math symbols
\usepackage{nicefrac}       % compact symbols for 1/2, etc.
\usepackage{microtype}      % microtypography
\usepackage{amsmath} 
\usepackage{graphicx}
\usepackage{caption}
\usepackage{subcaption}

\begin{document}

\title{Improved Particle Filters for Vehicle Localisation}

\author{\name Kira Kempinska \email kira.kowalska.13@ucl.ac.uk \\
       \addr Department of Security and Crime Science\\
       University College London\\
       London, WC1E 6BT
       \AND
       \name John Shawe-Taylor \email j.shawe-taylor@ucl.ac.uk \\
       \addr Department of Computer Science\\
       University College London\\
       London, WC1E 6BT\\}

%\editor{}

\maketitle
\begin{abstract}
The ability to track a moving vehicle is of crucial importance in numerous applications. The task has often been approached by the importance sampling technique of particle filters due to its ability to model non-linear and non-Gaussian dynamics, of which a vehicle travelling on a road network is a good example. Particle filters perform poorly when observations are highly informative. In this paper, we address this problem by proposing particle filters that sample around the most recent observation. The proposal leads to an order of magnitude improvement in accuracy and efficiency over conventional particle filters, especially when observations are infrequent but low-noise.

\end{abstract}

\section{Introduction}

Tracking a moving vehicle is a central and difficult problem arising in different contexts ranging from military applications to robotics \citep{Thrun2002,Gordon2002}. It consists of computing the best estimate of the vehicle's trajectory based on noisy sensor measurements. In this paper, we are interested in vehicle tracking when the road network is known.

Several strategies have been developed to track a vehicle on a road network \citep{Lou2009,Chawathe2007,Wenk2006,Alt2003,HuabeiYin,Pink2008}. We focus on the particle filter method \citep{Gordon1993}. The method has had numerous successes in this area due to its flexibility to handle cases where the dynamic and observation models are non-linear and/or non-Gaussian. It is an importance sampling technique that approximates the target distribution by sampling from a series of intermediate proposal distributions.  

Critically, in common with any important sampling method, the performance of particle filters is strongly dependent on the choice of the proposal distribution. If the proposal is not well matched to the target distribution, then the method produces samples that have low effective sample size and, as a result, it requires a prohibitively large number of particles to represent the target distribution accurately. The problem typically arises under highly informative observation regimes, in which the current observation provides significant information about the current state but the state dynamics are weak.  

The particle filter community has developed various approaches to mitigate the deficiency. One approach attaches a post-sampling step that moves particles sampled from the proposal distribution towards the target distribution using Markov Chain Monte Carlo moves \citep{Andrieu2010,Fox2001,VanDerMerwe2000} or by solving partial differential equations \citep{Li2016,Daum2010,Daum2007,Khan2014}. An alternative approach improves the proposal distribution by giving it additional information about the current \citep{Montemerlo2007,Kong1994,Liu1998} or even future observations \citep{Lin2013} or their approximations \citep{Pitt1999}. Several authors considered conditioning the proposal distribution on the current observation only \citep{Lin2005,Fox2001}. The approaches successfully increased the effective sample size, but, often at the cost of high computational complexity or analytical intractability. The construction of good, but also computationally efficient proposal distributions is still an open research question.

In this paper, we propose an improved particle sampling scheme that is both computationally efficient and mathematically robust. The approach generates proposals based on the current sensor observation only, leading to good alignment between the proposal and the target distribution even with a small sample size. It converges to the desired target distribution at faster rates than standard particle filters, especially when observations are highly informative, e.g. infrequent but low-noise. It is easy to implement and avoids the computational and analytical complexity of the discussed alternatives with other proposal distributions or post-sampling moves. It also presents a simpler approach to sample weighing than those previously proposed with the same proposal distribution \citep{Lin2005,Fox2001}.

The paper is structured as follows. We present the problem statement in Section \ref{problem_statement}, followed by a description of the standard particle filters in Section \ref{particle_filters}. We introduce the proposed particle filters method in Section \ref{improved_particle_filters}. We outline the application of the method to vehicle tracking in Section \ref{application} and present results in Section \ref{results}. We conclude by summarising the paper's contributions in Section \ref{conclusions}.

\section{Problem statement}  \label{problem_statement}

The key idea of particle filters is to estimate the marginal posterior distribution $p(x_{t} \mid z_{0:t})$ where $x_{t}$ is the state of the system at time $t$ and $z_{0:t} = \{z_0,\ldots,z_t\}$ is a sequence of measurements collected up to time step $t$. We call the posterior the \textit{belief} and use the following notation

\begin{equation}
Bel(x_t)=p(x_{t} \mid z_{0:t})
\label{eq:bel1}
\end{equation}

In the context of vehicle tracking, the belief is our estimate of the vehicle position at time $t$ given all measurements collected until then. The measurements include \textit{GPS readings} and \textit{controls}, which carry information about vehicle motion between consecutive timestamps. Denoting a GPS reading at time $t$ by $y_t$ and a control in the time interval $(t-1;t]$ by $u_{t-1}$, we have

\begin{equation}
Bel(x_t) = p(x_{t} \mid y_{0:t},u_{0:t-1})
\label{eq:bel2}
\end{equation}

Particle filters estimate $Bel(x_t)$ recursively. In order to arrive at a recursive equation, we note we can use Bayes rule to decompose Equation \ref{eq:bel2} to 

\begin{equation}
Bel(x_t)=\frac{p(y_t \mid x_t,y_{0:t-1},u_{0:t-1})p(x_t \mid y_{0:t-1},u_{0:t-1})}{p(y_t \mid y_{0:t-1},u_{0:t-1})}
\label{eq:bel3}
\end{equation}

The underlying assumption of particle filters is that the system follows the \textit{Markov assumption}, that is, measurements $y_t$ are conditionally independent of past measurements and controls given knowledge of the state $x_t$:

$$p(y_t \mid x_t,y_{0:t-1},u_{0:t-1}) = p(y_t \mid x_t)$$

This conveniently simplifies Equation \ref{eq:bel3} to 

$$Bel(x_t)=\frac{p(y_t \mid x_t)p(x_t \mid y_{0:t-1},u_{0:t-1})}{p(y_t \mid y_{0:t-1},u_{0:t-1})}$$

We integrate out the position at $x_{t-1}$ in order to arrive at the following recursive form

$$Bel(x_t)=\frac{p(y_t \mid x_t)}{p(y_t \mid y_{0:t-1},u_{0:t-1})}\int p(x_t \mid x_{t-1},y_{0:t-1},u_{0:t-1})p(x_{t-1} \mid y_{0:t-1},u_{0:t-1}) dx_{t-1}$$

which can be simplified again using \textit{Markov assumption} by noting that:

$$p(x_t \mid x_{t-1},y_{0:t-1},u_{0:t-1})=p(x_t \mid x_{t-1},u_{t-1})$$ 

Finally, we arrive at a recursive estimator known as \textit{Bayes filter}:

\begin{align}
\begin{split}
Bel(x_t)= & \frac{p(y_t \mid x_t)}{p(y_t \mid y_{0:t-1},u_{0:t-1})}\int p(x_t \mid x_{t-1},u_{t-1})Bel(x_{t-1}) dx_{t-1} \\
		= & \,\eta\,\,p(y_t \mid x_t)\int p(x_t \mid x_{t-1},u_{t-1})Bel(x_{t-1}) dx_{t-1}
\end{split}
\label{eq:bf}
\end{align}

where $\eta$ is a normalising constant. The \textit{Bayes filter} equation is the basis for particle filters and the improved particle filters that we propose in this paper.

\section{Particle Filters}  \label{particle_filters}

Particle filters approximate the belief $Bel(x)$ by a set of $m$ weighted samples distributed according to $Bel(x)$:

$$Bel(x)=\{x^{(i)},w^{(i)}\}_{i=1,\ldots,m}$$

where each $x^{(i)}$ is a sample (a state) and $w^{(i)}$ are non-negative weights called \textit{importance factors} that determine the importance of each sample.

The particle filters method operates recursively. It begins by generating $m$ samples $x_0^{(i)}$ from the initialisation distribution $Bel(x_0) = p(x_0)$ and annotates them by the uniform importance factor $1/m$.  Subsequently, it estimates $Bel(x_t)$ at any future timestamp $t$ by performing a three-step recursive update, computing the expression in Equation \ref{eq:bf} \textit{from the right to the left}.

\vspace{3mm} 

for $k=1,\ldots,m:$
\begin{enumerate}
	\item Sample a state $x_{t-1}$ by drawing a random $x_{t-1}^{(i)}$ from the sample set representing $Bel(x_{t-1})$ according to the distribution defined through the importance factors $w_{t-1}^{(i)}$.
	\item Use the sample $x_{t-1}^{(i)}$ and the control $u_{t-1}$ to generate a sample $x_{t}^{(j)}$ according to the so-called \textit{transition probability} $p(x_t \mid x_{t-1},u_{t-1})$.
	\item Finally, use the observation $y_t$ to weigh the sample $x_{t}^{(j)}$ by the non-normalized importance factor given by the so-called \textit{observation probability} $p(y_t \mid x_{t}^{(j)})$, the likelihood of the sample $x_{t}^{(j)}$ given the observation $y_t$.
\end{enumerate}

\vspace{3mm} 

Further below, it will be important to notice that the particle filters method is, in fact, an importance sampling scheme. It approximates $Bel(x_t)$ using a proposal distribution given by

\begin{equation}
Q = p(x_t \mid x_{t-1},u_{t-1})Bel(x_{t-1})
\label{eq:q_pf}
\end{equation}

The proposal approximates the desired posterior

\begin{equation}
P = \frac{p(y_t \mid x_t) p(x_t \mid x_{t-1},u_{t-1})Bel(x_{t-1})}{p(y_t \mid y_{0:t-1},u_{0:t-1})} 
\end{equation}
\begin{samepage}
Consequently, the importance factors are given by the quotient

\begin{align}
\begin{split}
\frac{P}{Q}= & \,\,[p(x_t \mid x_{t-1},u_{t-1})Bel(x_{t-1})]^{-1}\frac{p(y_t \mid x_t) p(x_t \mid x_{t-1},u_{t-1})Bel(x_{t-1})}{p(y_t \mid y_{0:t-1},u_{0:t-1})}  \\
			\propto & \,\,p(y_t \mid x_t)
\end{split}
\label{eq:if_pf}
\end{align}
\end{samepage}

\section{Improved Particle Filters}  \label{improved_particle_filters}

We propose an improved particle sampling scheme in which $x_t$ are sampled directly around the most recent observation $y_t$ according to the proposal distribution:

\begin{equation}
Q_{new}= \frac{p(y_t \mid x_t)}{\pi(y_t)}  \qquad\text{with}\qquad 	\pi(y_t )=\int p(y_t \mid x_t)dx_t  
\label{eq:q_new}
\end{equation} 

This new proposal distribution possesses orthogonal strengths to the one in Equation \ref{eq:q_pf}, in that it generates samples that are highly consistent with the most recent sensor measurement but ignorant of past measurements and controls. As such, we expect it to outperform conventional particle filters in systems where the current observation provides more information about the current state than the underlying state dynamics. 

The importance factors for these samples are again calculated by the quotient:

\begin{align}
\begin{split}
\frac{P}{Q_{new}}=&\left[\frac{p(y_t \mid x_t)}{\pi(y_t)}\right]^{-1}\frac{p(y_t \mid x_t) p(x_t \mid x_{t-1},u_{t-1})Bel(x_{t-1})}{p(y_t \mid y_{0:t-1},u_{0:t-1})}  \\ \\
				=& \,\,\frac{p(x_t \mid x_{t-1},u_{t-1})Bel(x_{t-1}) \pi(y_t) }{p(y_t \mid y_{0:t-1},u_{0:t-1})}  \\ \\
				\propto & \,\,p(x_t \mid x_{t-1},u_{t-1})Bel(x_{t-1})	
\end{split}
\end{align} 

Since $Bel(x_{t-1})$ is represented by a set of samples $x_{t-1}^{(i)}$ weighted by importance factors $w_{t-1}^{(i)}$, the (non-normalised) importance factor for any sample $x_t^{(j)}$ can be approximated by

\begin{equation}
\sum_{i=1}^m p(x_t^{(j)} \mid x_{t-1}^{(i)},u_{t-1})w_{t-1}^{(i)}
\label{eq:if_new}
\end{equation}

The importance factor reflects the likelihood of the sample given \textit{past} measurements and controls. This is orthogonal to the previous definition in Equation \ref{eq:if_pf}, where it depends on the \textit{current} measurement only.

Overall, the proposed sampling scheme changes how data and controls are used in belief estimation: the current measurement is now used for sampling (instead of weighing); past measurements and controls are used for calculating importance factors (instead of sampling). 

The scheme is implemented recursively. It initialises $Bel(x_0)$ by generating $m$ samples around the first observation $y_0$ according to the observation probability $p(y_t \mid x_t)$. The samples are assigned the uniform importance factor of $1/m$. Subsequently, it estimates $Bel(x_t)$ at  timestamps $t>0$ using a two-step recursive update:
 
\begin{samepage}
for $k=1,\ldots,m:$
\begin{enumerate}
	\item Generate a sample $x_t^{(i)}$ according to the observation probability $p(y_t \mid x_t)$.
	\item Use the sample set representing $Bel(x_{t-1})$ to weight the sample $x_t^{(i)}$ by the importance factor in Equation \ref{eq:if_new} , the likelihood of the sample $x_t^{(i)}$ given past measurements and controls.
\end{enumerate}
\end{samepage}
\vspace{3mm} 

In the context of vehicle tracking, the estimates of $Bel(x_t)$ approximate the vehicle \textit{position} at time $t$. If instead of the single-time approximation, you are interested in finding the most likely \textit{trajectory} that the vehicle traversed until time $t$, it can be computed via the following dynamic programming routine. It corresponds to finding the sequence $x_{0:t}$ that maximises the posterior $p(x_{0:t} \mid y_{0:t}, u_{0:t-1})$.
\begin{enumerate}
	\item Choose a sample $x_t^{(i)}$ from the sample set representing $Bel(x_t)$ that has the highest importance factor $w_t^{(i)}$.	
	\item Use the sample $x_t^{(i)}$ to find a preceding sample $x_{t-1}^{(j)}$ from $Bel(x_{t-1})$ that maximises $p(x_t^{(i)} \mid x_{t-1}^{(j)},u_{t-1})w_{t-1}^{(j)}$, i.e. is the most likely preceding state. Repeat this step until you reach $t=0$.
\end{enumerate}

\section{Application to Vehicle Tracking}  \label{application}

\subsection{Data}

We tested the improved particle filters method on a GPS trajectory of a police patrol vehicle during its night shift (9am to 7am) in the London Borough of Camden on February $9^{th}$ 2015. The dataset contains 4,800 GPS points that were emitted roughly every second when moving. It was acquired for research purposes as part of the "Crime, Policing and Citizenship" project\footnote{UCL Crime Policing and Citizenship: \url{http://www.ucl.ac.uk/cpc/}}. 

\subsection{Implementation}

In order to apply the improved particle filters to vehicle tracking, we need to specify the form of the observation probability $p(y_t \mid x_t)$ and the transition probability $p(x_t \mid x_{t-1},u_{t-1})$. Their forms depend on the vehicle's dynamics and the type of sensor used for localisation (a GPS receiver in our case). The distributions are time-invariant; hence we will omit the time index $t$ in the following derivations. 

\subsubsection{Observation probability}

\paragraph{Definition}

We model the conditional probability $p(y \mid x)$ of observing a GPS point $\boldsymbol{y}$, represented by its easting and northing coordinates:

\begin{equation}
\boldsymbol{y} = \left(\begin{array}{cc} y_{e} \\ y_{n}  \end{array}\right)
\label{eq:y_t}
\end{equation}

as a two-dimensional Gaussian distribution

$$\boldsymbol{y} \sim \mathcal{N} (\boldsymbol{\mu},\boldsymbol{\Sigma})$$

with the mean vector $\boldsymbol{\mu}$ representing the true vehicle position $\boldsymbol{x}$

\begin{equation}
\boldsymbol{\mu}=\boldsymbol{x}=\left(\begin{array}{cc} x_{e} \\ x_{n}  \end{array}\right)
\label{eq:mu}
\end{equation}

and the covariance $\boldsymbol{\Sigma}$ that is constant across space, i.e. \textit{isotropic} covariance 

\begin{equation}
\boldsymbol{\Sigma} = \left[\begin{array}{cc} \Sigma_{ee} & \Sigma_{en} \\ \Sigma_{ne} & \Sigma_{nn} \end{array}\right] =\left[\begin{array}{cc} \sigma^2 & 0 \\ 0 & \sigma^2 \end{array}\right] 
\label{eq:cov}
\end{equation}

This representation of $p(y \mid x)$ reflects our expectation that GPS observations are normally distributed around the true vehicle positions.

\paragraph{Proposal Generation}

\begin{figure}
    \centering
    \begin{subfigure}[b]{0.45\textwidth}
        \includegraphics[width=\textwidth]{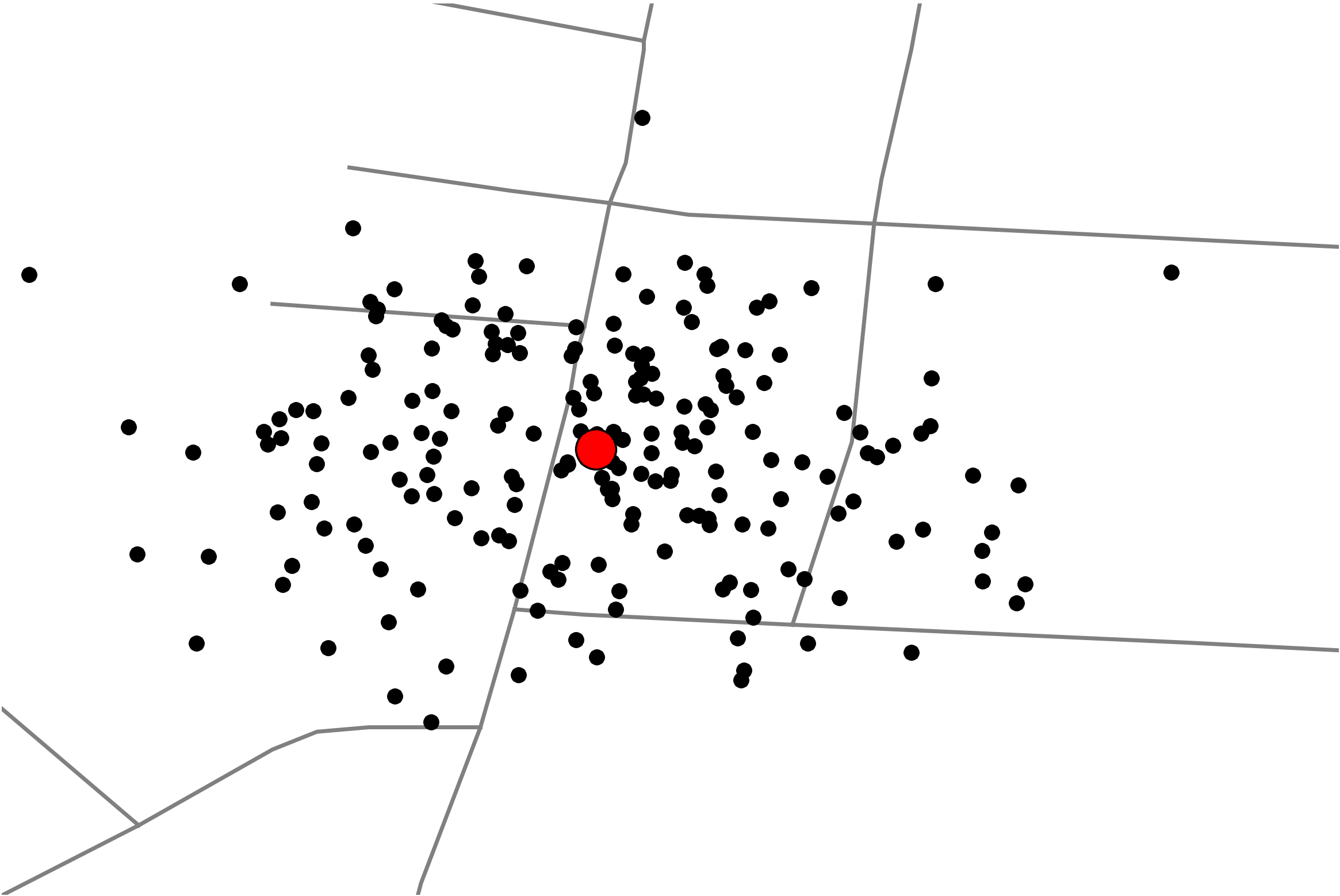}
        \caption{unconstrained}
        \label{fig:pf_init_space}
    \end{subfigure}
    \hspace{2em}
    %~
    %~ %add desired spacing between images, e. g. ~, \quad, \qquad, \hfill etc. 
      %(or a blank line to force the subfigure onto a new line)
    \begin{subfigure}[b]{0.45\textwidth}
        \includegraphics[width=\textwidth]{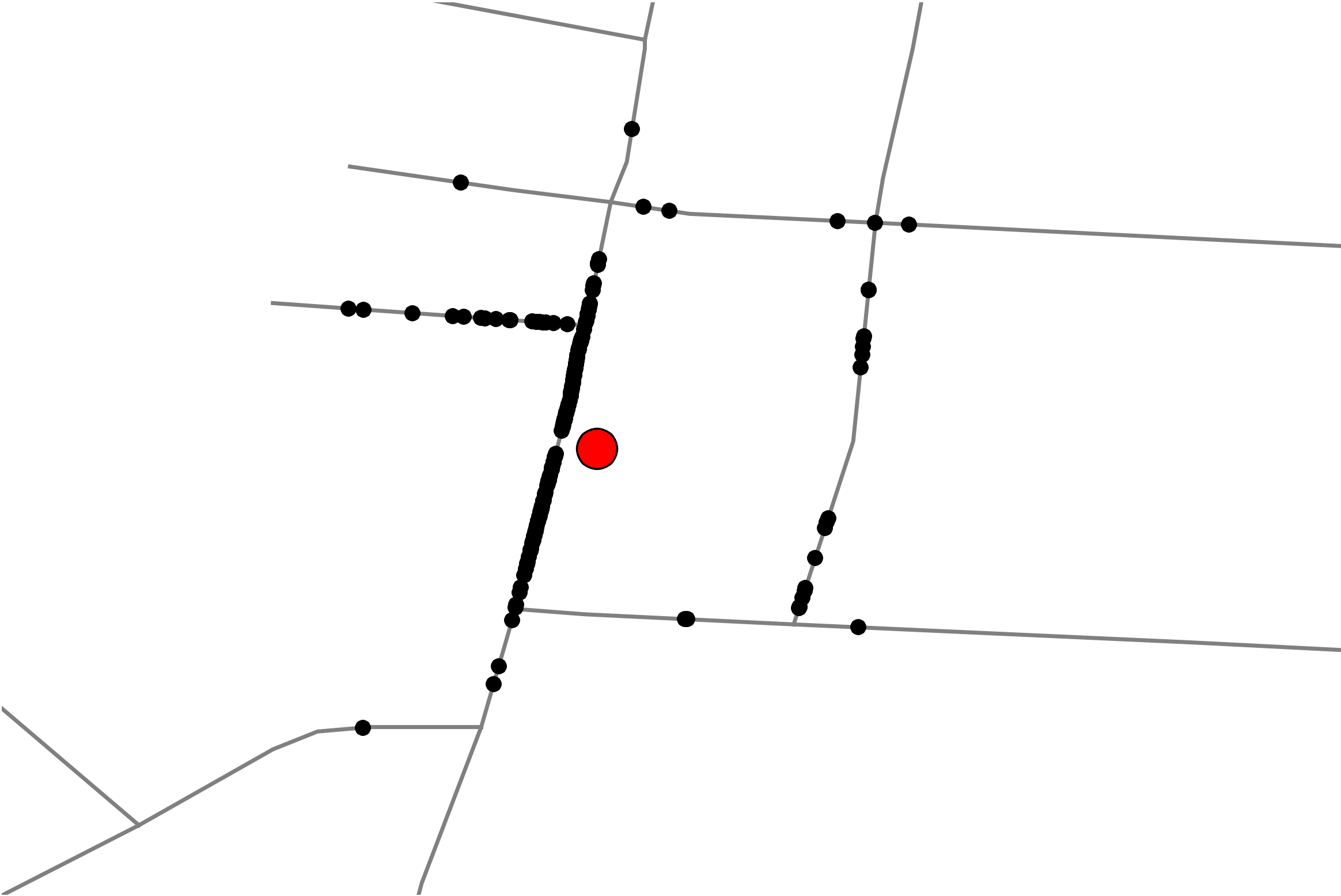}
        \caption{constrained to the road network}
        \label{fig:pf_init_network}
    \end{subfigure}
    \caption{Sampling proposal positions around a GPS point (red).}\label{pf_init}
\end{figure}

In the proposed method, we use the observation probability $p(y \mid x)$ to sample possible vehicle positions on the road network (see Equation \ref{eq:q_new}). In order to efficiently generate samples on the road network (as shown in Figure \ref{fig:pf_init_network}), we want to analytically project the two-dimensional $p(y \mid x)$ onto individual road segments. 

%\begin{figure}[h]
%  \centering
%  \fbox{\includegraphics[width=0.5\textwidth]{figures/gaussian_over_network.png}}
%  \caption{The observation probability $p(y_t \mid x_t)$ (blue) around the true vehicle position $x_t$ (red).}
%  \label{fig:gaussian}
%\end{figure}

We begin with the general form of a two-dimensional Gaussian distribution for $p(y \mid x)$

\begin{equation}
p(y \mid x) = \mathcal{N} (\boldsymbol{y} \mid \boldsymbol{\mu},\boldsymbol{\Sigma})=\frac{1}{2\pi|\boldsymbol{\Sigma}|^{1/2}}exp\left \{-\frac{1}{2}(\boldsymbol{y}-\boldsymbol{\mu})^T\boldsymbol{\Sigma}^{-1}(\boldsymbol{y}-\boldsymbol{\mu})\right \}
\label{eq:p_y_given_x1}
\end{equation}

We precompute the inverse of the covariance matrix

$$\boldsymbol{\Sigma^{-1}}=\frac{1}{\sigma^4}\left[\begin{array}{cc} \sigma^2 & 0 \\ 0 & \sigma^2 \end{array}\right]=\left[\begin{array}{cc} \sigma^{-2} & 0 \\ 0 & \sigma^{-2} \end{array}\right]$$

and use it together with the partitioning (\ref{eq:y_t}), (\ref{eq:mu}), and (\ref{eq:cov}) to rewrite (\ref{eq:p_y_given_x1}) as

\begin{align}
\begin{split}
p(y \mid x) =& \,\,\frac{1}{2\pi\sigma^2}exp\left \{-\frac{1}{2}\left[\frac{(y_e-\mu_e)^2}{\sigma^2}+\frac{(y_n-\mu_n)^2}{\sigma^2}\right]\right \} \\
				=& \,\,\frac{1}{(2\pi\sigma^2)^{1/2}}exp\left \{-\frac{1}{2\sigma^2}(y_e-\mu_e)^2\right \} \times \frac{1}{(2\pi\sigma^2)^{1/2}}exp\left \{-\frac{1}{2\sigma^2}(y_n-\mu_n)^2\right \} \\ \\
				=& \,\,\mathcal{N} (y_e \mid \mu_e,\sigma) \times \mathcal{N} (y_n \mid \mu_n,\sigma)
\end{split}
\end{align}

We successfully factor $p(y \mid x)$ into a product of two Gaussian distributions along the \textit{easting} and \textit{northing} directions due to the isotropic properties of the covariance matrix in (\ref{eq:cov}). In fact, the factorisation of $p(y \mid x)$ holds for any other orthogonal coordinate system. Therefore, we replace the easting-nothing coordinates with orthogonal distances from $x$ dictated by the road segment that $x$ is on: \textit{$a$} (distance \textit{to} the road segment), \textit{$b$} (distance \textit{along} the road segment).

Under the new coordinate system $x$ and $y$ are partitioned as
$$x=\left(\begin{array}{cc} x_{a} \\ x_{b}  \end{array}\right)=\left(\begin{array}{cc} 0 \\ 0  \end{array}\right) \qquad\qquad y = \left(\begin{array}{cc} y_{a} \\ y_{b}  \end{array}\right)$$
and $p(y \mid x)$ becomes
\begin{align}
\begin{split}
p(y \mid x) =& \mathcal{N} (y_{a} \mid \mu_{a},\sigma) \times \mathcal{N} (y_{b} \mid \mu_{b},\sigma) \\
				=& \mathcal{N} (y_{a} \mid 0,\sigma) \times \mathcal{N} (y_{b} \mid 0,\sigma)
\end{split}
\label{eq:p_y_t_new}
\end{align}
The above definition enables us to generate proposals $x$ in accordance with the observation model $p(y \mid x)$ (as specified in Equation \ref{eq:q_new}):
\begin{enumerate}
	\item Firstly, sampling a road segment that $x$ in on such that $y_{a}$ $\sim \mathcal{N} (0,\sigma)$
	\item Secondly, sampling the position of $x$ along the segment such that $y_{b}$  $\sim \mathcal{N} (0,\sigma)$
\end{enumerate}

\subsubsection{Transition probability}

We set the transition probability $p(x_t \mid x_{t-1},u_{t-1})$ to be a linear estimate equal to the Cartesian distance between GPS points $x_{t-1}$ and $x_t$ (the control $u_t$) plus an additive Gaussian noise. This is a simplistic assumptions that could be further explored, however, it is not the focus of this paper.

\subsection{Validation}
In the absence of the ground truth about vehicle positions at any point in time, we propose a validation framework based on the well-established technique of cross-validation \citep{Barber2012}. We remove every 10th GPS points from the available GPS trajectory. We then infer the path taken by the vehicle given the incomplete trajectory and the road network. Finally, we measure the distance between each removed point and the inferred path. The distances across all removed points form the distribution of the prediction error.

%\begin{figure}[htbp] \begin{center} 
%\resizebox{0.5\textwidth}{!}{ 
%	\includegraphics{figures/crossvalidation_labelled.png}
%} \caption{Exemplary GPS trajectory with points split into training and test sets.} \label{fig:crossvalidation} \end{center} \end{figure} %

\section{Results}  \label{results}

\begin{figure}
    \centering
    \begin{subfigure}[b]{0.48\textwidth}
        \includegraphics[width=\textwidth]{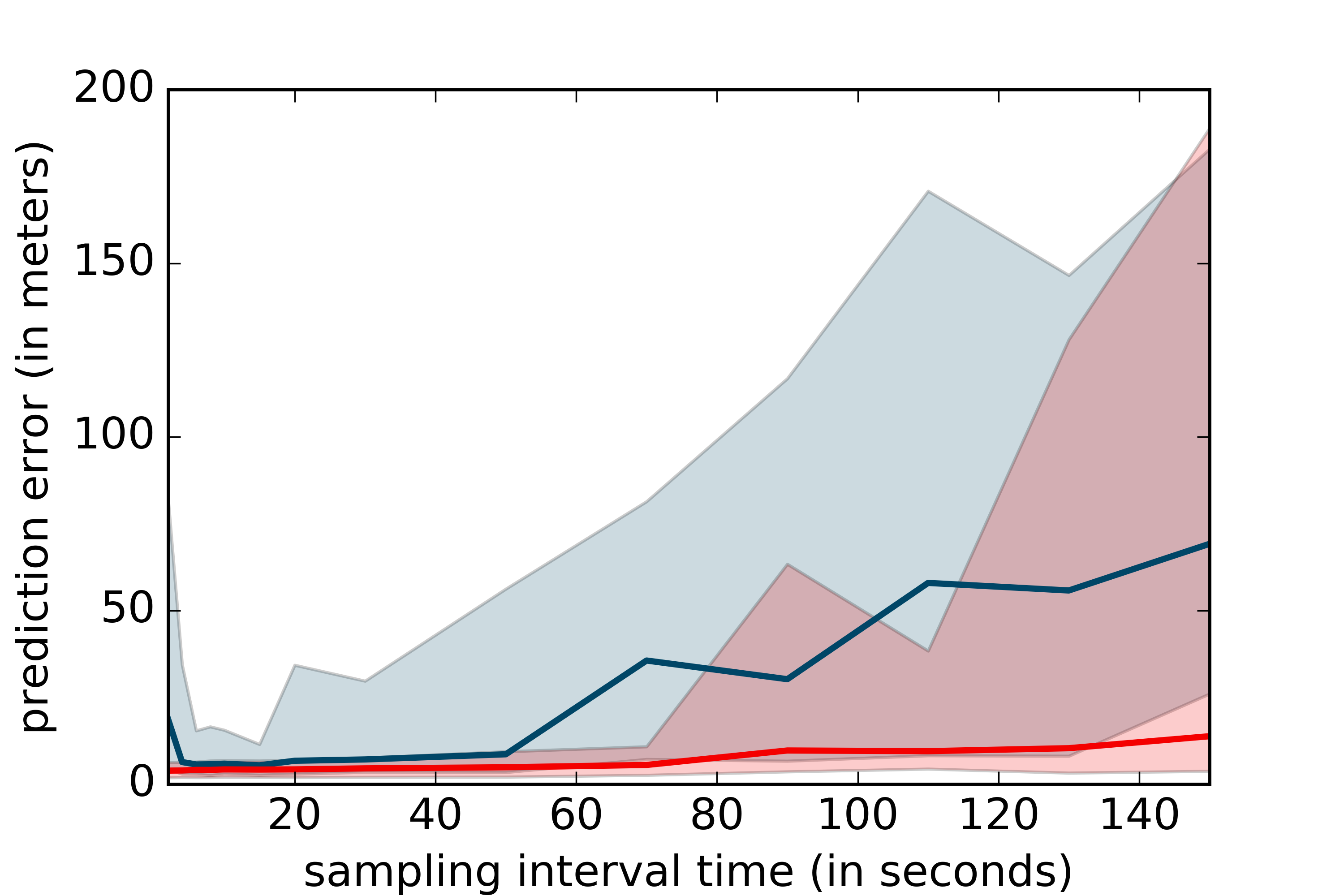}
        \caption{sampling rate}
        \label{fig:accuract_sr}
    \end{subfigure}
    \hspace{1em}
    %~
    %~ %add desired spacing between images, e. g. ~, \quad, \qquad, \hfill etc. 
      %(or a blank line to force the subfigure onto a new line)
    \begin{subfigure}[b]{0.48\textwidth}
        \includegraphics[width=\textwidth]{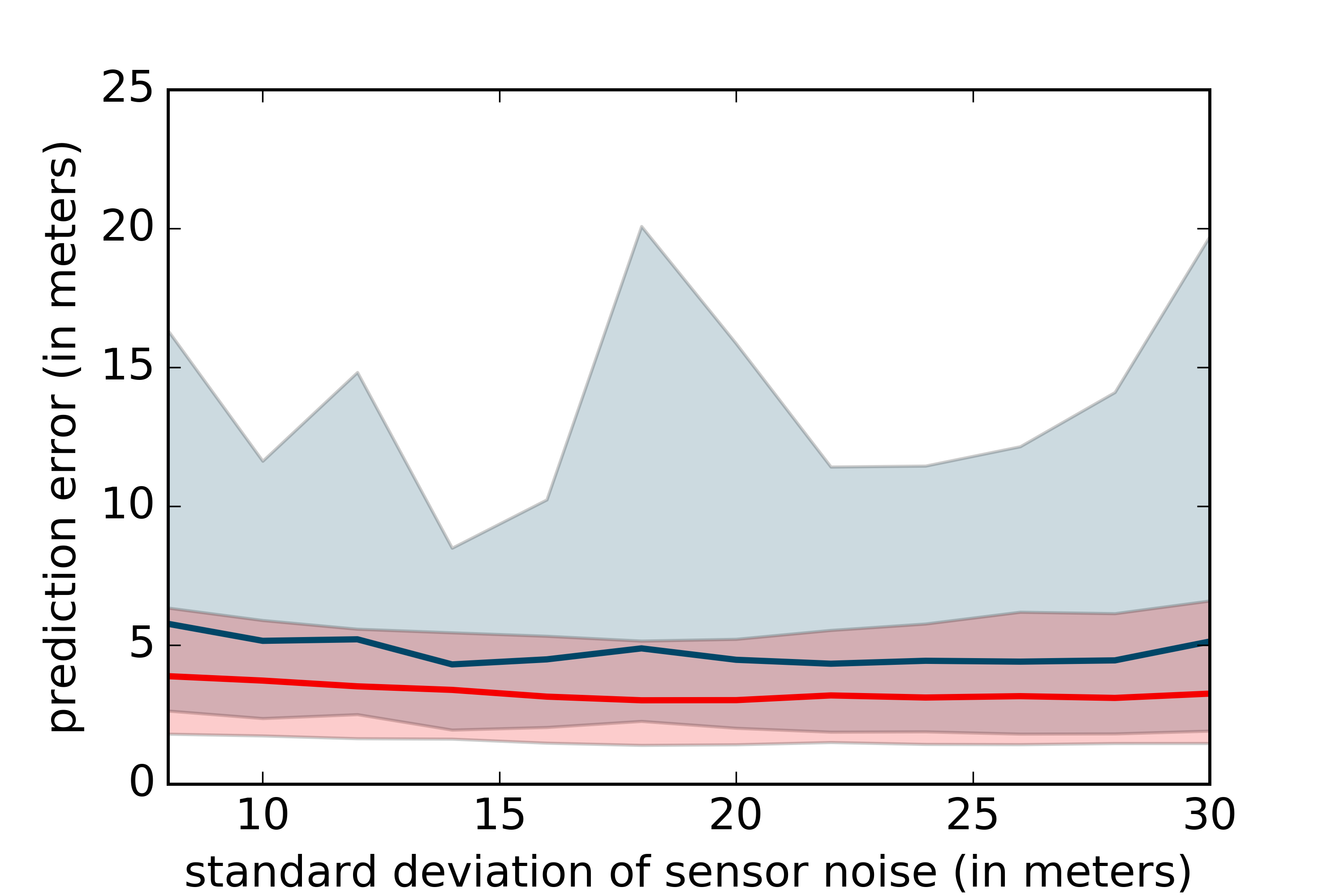}
        \caption{sensor noise}
        \label{fig:accuracy_me}
    \end{subfigure}
    \caption{Accuracy of the improved particle filters (red) and the standard particle filters (blue) on GPS data with varied sampling rate and sensor noise, represented as 25th, 50th and 75th percentiles of prediction errors.}\label{fig:pf_pfla_accuracy}
\end{figure}

A series of tests was conducted to elucidate the difference between the standard and the proposed particle filters. We found that the modified proposal distribution consistently outperforms conventional particle filters in terms of accuracy. As expected, largest gains in accuracy are observed on datasets with long sampling intervals  as their observations are infrequent and hence become highly informative. Figure \ref{fig:pf_pfla_accuracy} plots the prediction error (in meters) of both algorithms for different sampling intervals and levels of sensor noise, using $m=10$ samples only. It shows that the proposed method has lower \textit{median} error across all examined sampling rates and sensor noise levels, as well as much lower error \textit{variation}. 

\begin{figure}
    \centering
    \begin{subfigure}[b]{0.48\textwidth}
        \includegraphics[width=\textwidth]{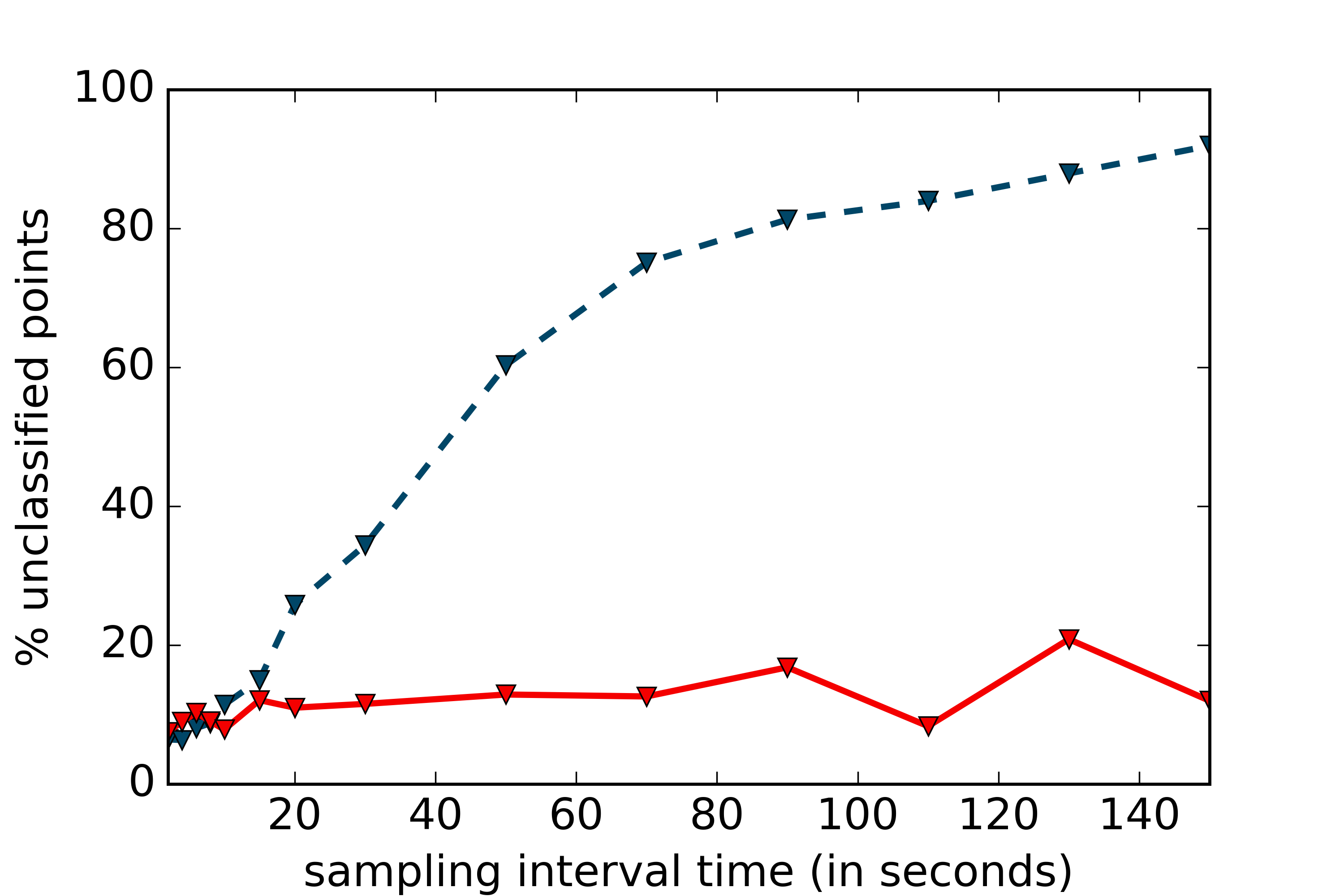}
        \caption{sampling rate}
        \label{fig:pclass_sr}
    \end{subfigure}
    \hspace{1em}
    %~
    %~ %add desired spacing between images, e. g. ~, \quad, \qquad, \hfill etc. 
      %(or a blank line to force the subfigure onto a new line)
    \begin{subfigure}[b]{0.48\textwidth}
        \includegraphics[width=\textwidth]{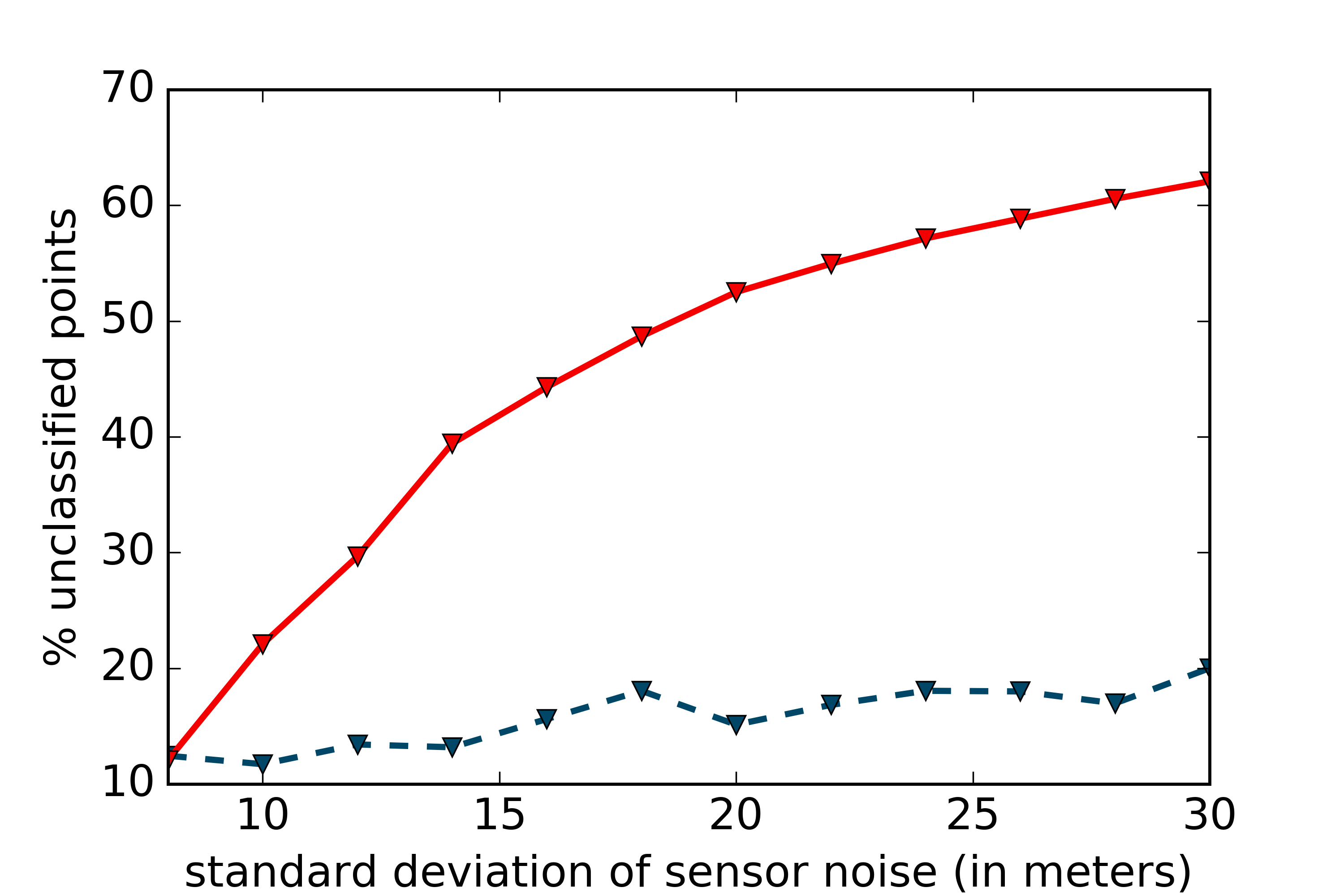}
        \caption{sensor noise}
        \label{fig:pclass_me}
    \end{subfigure}
    \caption{Percentage of time the improved particle filters (red) and the conventional particle filters (blue) lost track of the position of the vehicle as a function of the GPS sampling rate and the sensor noise.}\label{fig:pf_st_pclass}
\end{figure}

We evaluated the ability of both methods to track a vehicle over time. When they fail to track a vehicle, it means that all positions that they propose are completely unlikely given sensor data, i.e. particle weights sum up to zero. The standard particle filter basically fails when sensor measurements are infrequent (with $m=10$ samples). Figure \ref{fig:pclass_sr} shows that it is unable to track the vehicle nearly 70\% of the time when the sampling interval increases to one minute. In the same scenario, the proposed method gives excellent results that show little variation to changes to sampling intervals. 

On the contrary, the proposed method fails to track when sensors are very noisy. Although it shows high accuracy (see Figure \ref{fig:accuracy_me}), it is prone to high failure rates as the level of sensor noise increases (Figure \ref{fig:pclass_me}). This weakness reflects the orthogonal limitations of the two approaches: our method generates samples that are highly consistent with the most recent measurement (which makes it sensitive to sensor noise), whereas the conventional approach samples in accordance with past measurements (inefficient when sampling rates are low).

\begin{figure}
    \centering
    \begin{subfigure}[b]{0.48\textwidth}
        \includegraphics[width=\textwidth]{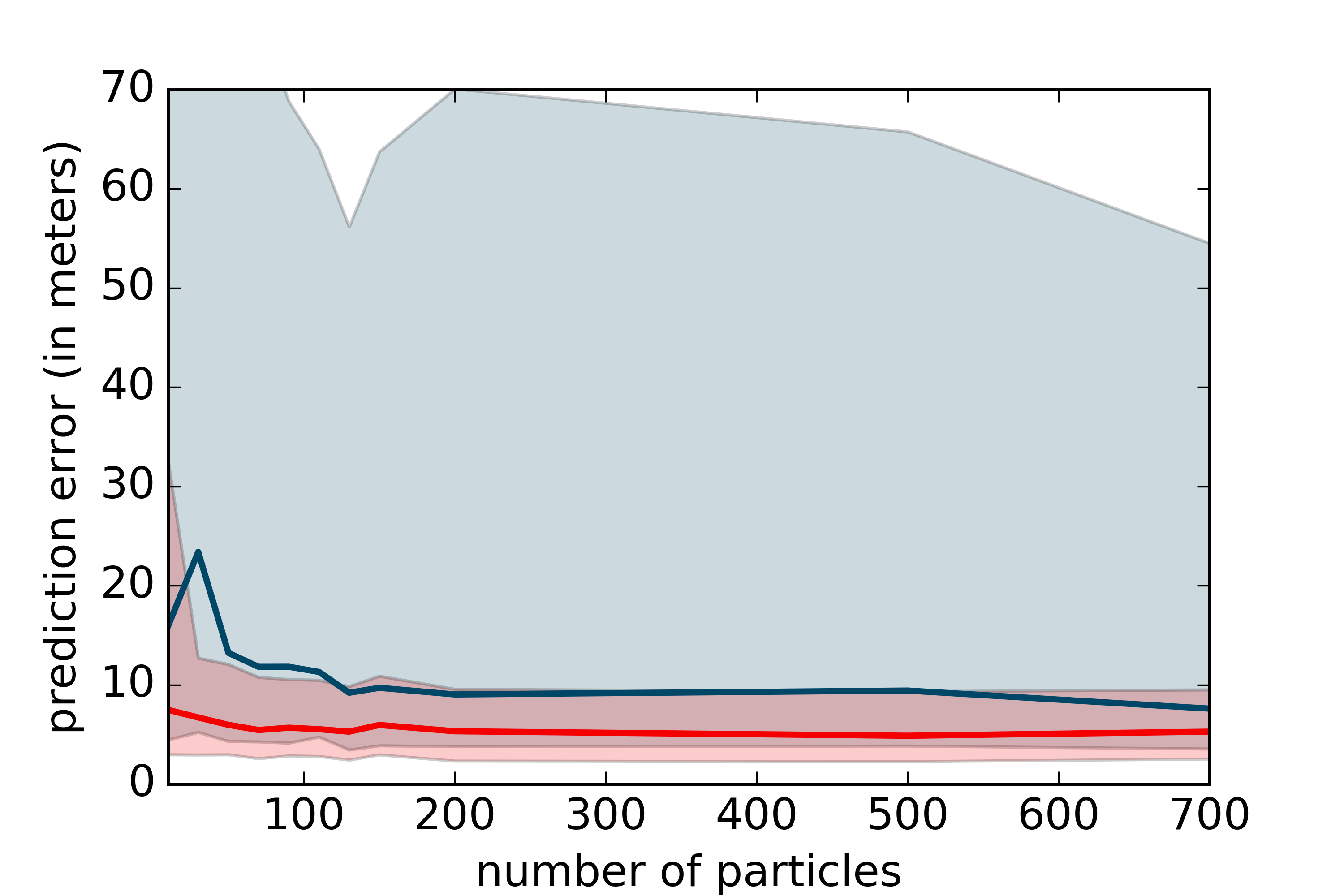}
        \caption{accuracy}
    \end{subfigure}
    \hspace{1em}
    %~
    %~ %add desired spacing between images, e. g. ~, \quad, \qquad, \hfill etc. 
      %(or a blank line to force the subfigure onto a new line)
    \begin{subfigure}[b]{0.48\textwidth}
        \includegraphics[width=\textwidth]{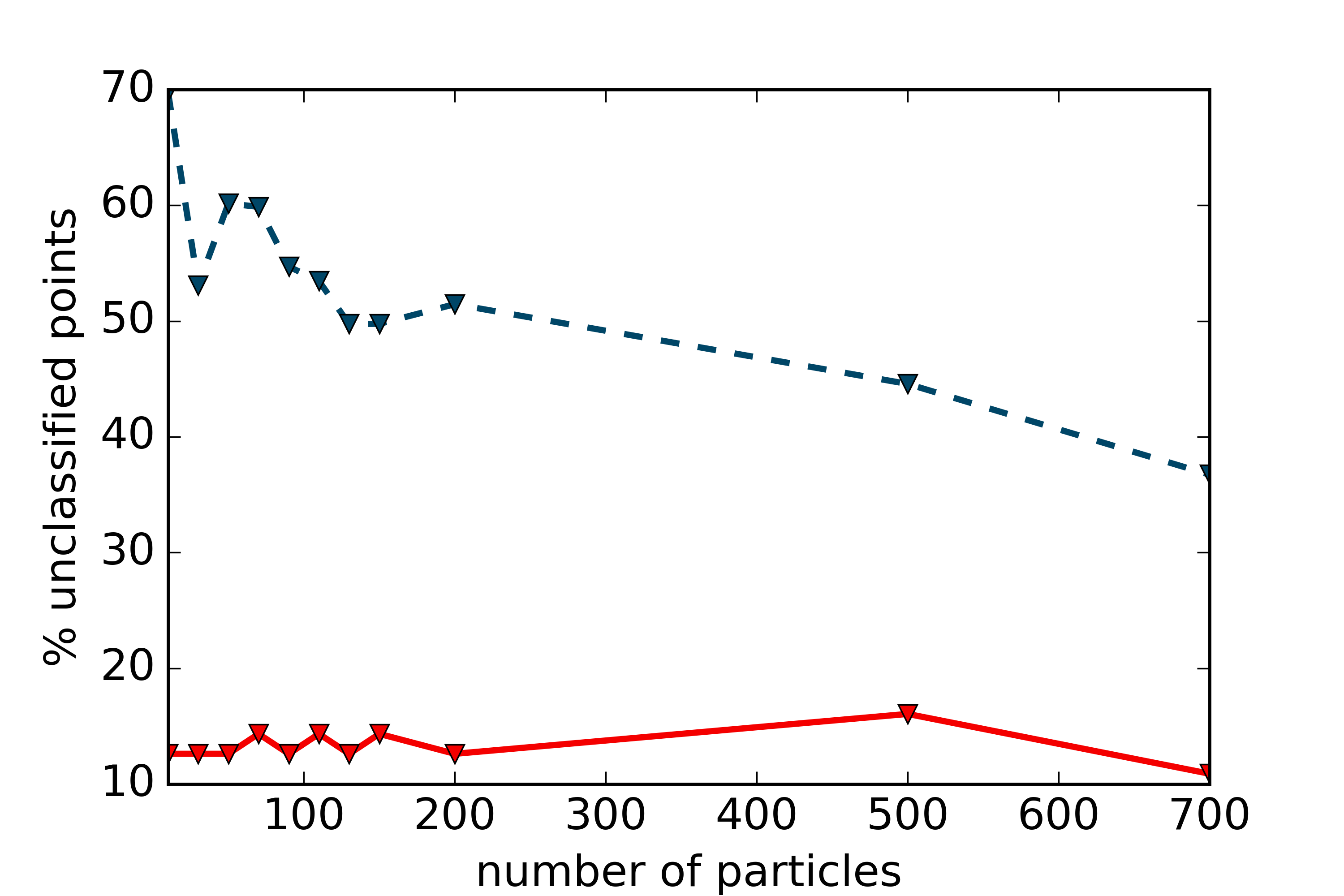}
        \caption{failure rate}
    \end{subfigure}
    \caption{Accuracy and robustness of the improved particle filters (red) and the conventional particle filters (blue) as a function of the number of samples used. Accuracy is shown as the 25th, 50th and 75th percentiles of prediction errors.}\label{fig:pf_st_samples}
\end{figure}

\begin{figure}
    \centering
    \begin{subfigure}[b]{0.48\textwidth}
        \includegraphics[width=\textwidth]{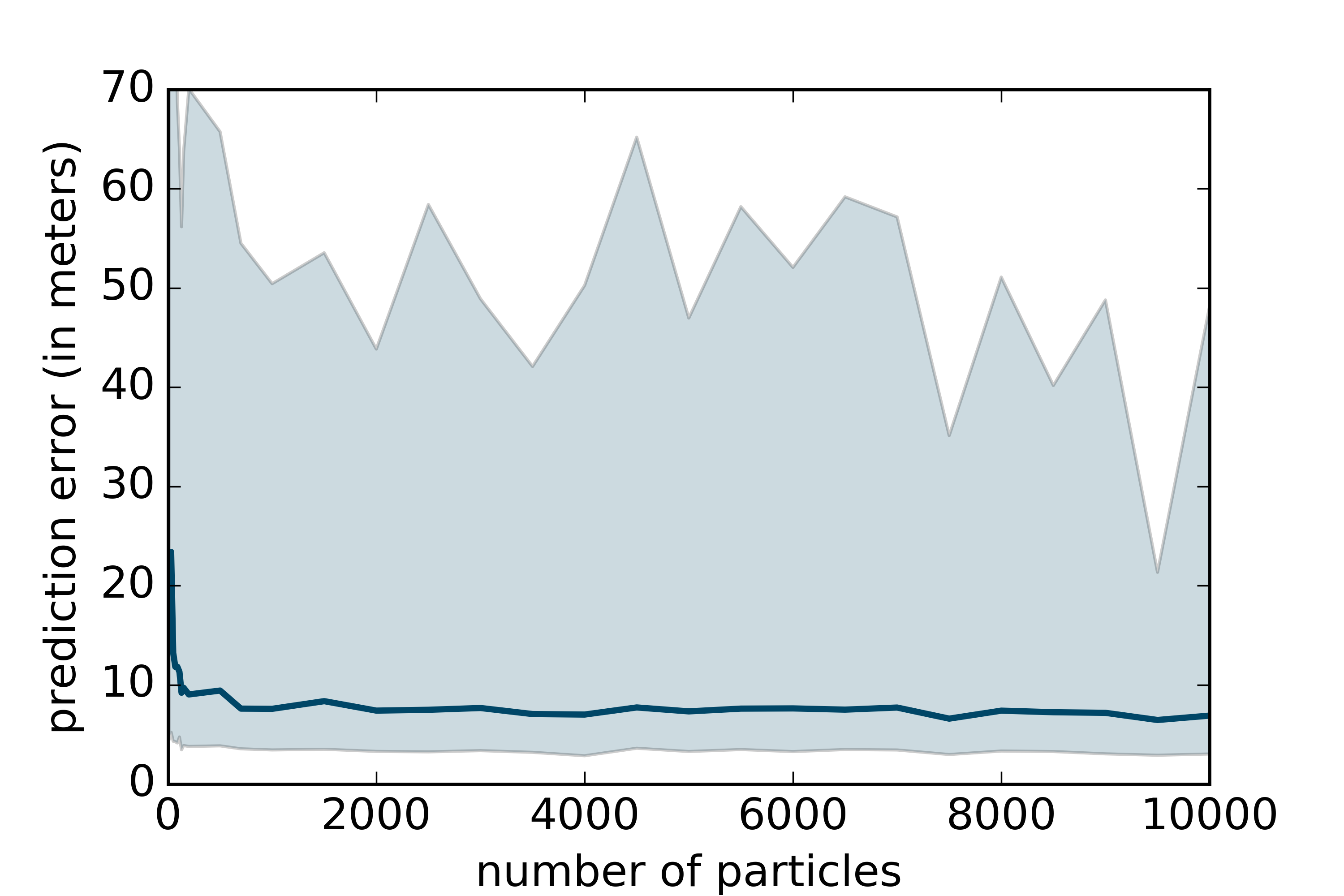}
        \caption{accuracy}
    \end{subfigure}
    \hspace{1em}
    %~
    %~ %add desired spacing between images, e. g. ~, \quad, \qquad, \hfill etc. 
      %(or a blank line to force the subfigure onto a new line)
    \begin{subfigure}[b]{0.48\textwidth}
        \includegraphics[width=\textwidth]{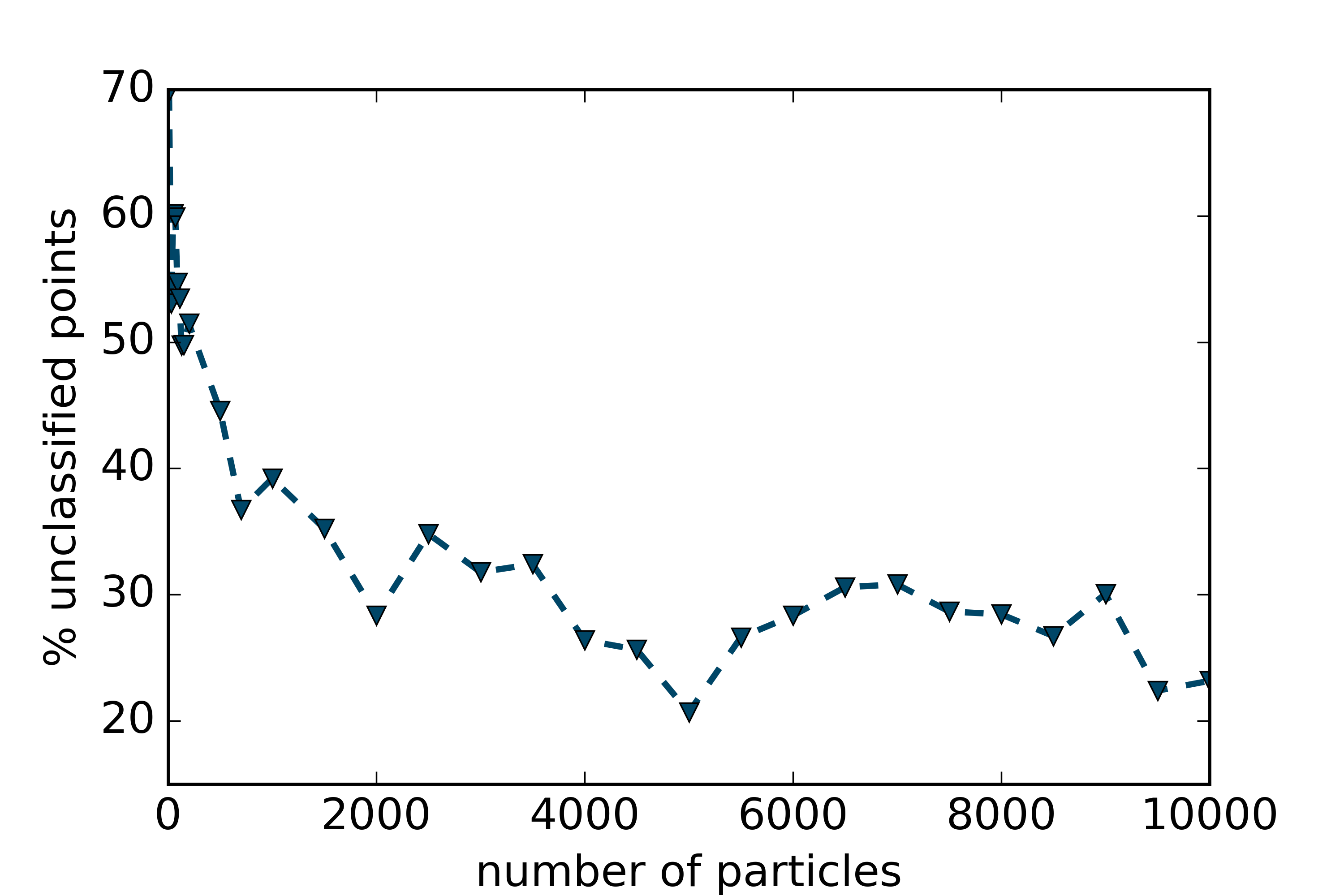}
        \caption{failure rate}
    \end{subfigure}
    \caption{Accuracy and robustness of the standard particle filters as the number of samples is increased to very large values. Accuracy is shown as the 25th, 50th and 75th percentiles of prediction errors.}\label{fig:pf_longterm}
\end{figure}

Finally, we tested the sensitivity of the proposed method to the number of samples used. Figure \ref{fig:pf_st_samples} shows comparative results on GPS data with the sampling interval of 70 seconds. The proposed method yields significantly better results, both in terms of accuracy and robustness to failure. When only $m=10$ samples are used, it reduces the estimation error by almost 10 meters and the percentage of failure by as much as 68\%. The performance is further improved when more samples are used, but the gain is small compared to the conventional particle filters. In fact, the proposed method with $m=100$ samples is more accurate and robust than the conventional particle filters with as many as $m=10000$ samples (see Figure \ref{fig:pf_longterm}). Therefore, it can be reliably used with a small number of samples, making it highly computationally efficient. 

\begin{samepage}
\section{Conclusions}  \label{conclusions}

This paper describes a modified particle filters method that shows uniformly superior accuracy to the conventional particle filters. The improved algorithm utilizes a different proposal distribution which uses only the most recent observation in the position prediction process. In doing so, it makes more efficient use of the particles, particularly in situations in which the transition noise is high in relation to the observation noise.

The main contribution of the paper is the proposal distribution itself and the derivation of the associated importance weights that guarantees convergence to the same posterior distribution as the standard particle filters. An important contribution is also the projection of a two-dimensional Gaussian onto a network of roads, which enables efficient sampling on the road network from a spatial Gaussian.

The theoretical contributions are complemented by experimental results of vehicle tracking using a police GPS dataset. The new algorithm is consistently more accurate than the standard particle filters, with largest gains in accuracy on sparse GPS data. It requires much fewer samples to yield good performance. In fact, as few as fifty samples are sufficient to outperform the standard method with 10,000 particles in terms of accuracy and proneness to failure. We believe that our results illustrate that particle filters can be radically improved if one carefully chooses a proposal distribution, such that it extracts the most information from the available data.
\end{samepage}
\section*{Acknowledgements}
This work is part of the project - Crime, Policing and Citizenship (CPC): Space-Time Interactions of Dynamic Networks (www.ucl.ac.uk/cpc), supported by the UK Engineering and Physical Sciences Research Council (EP/J004197/1). The data provided by Metropolitan Police Service (London) is greatly appreciated.

We would also like to show our gratitude to Dr Simon Julier for very helpful discussions during the course of this research.

\bibliography{references}

\end{document}